\documentclass{article}

\usepackage{spconf,amsmath,graphicx}
\usepackage{booktabs}
\usepackage{multirow}
\usepackage{url}
\usepackage{hyperref}

\title{Mask-based Invisible Backdoor Attacks on Object Detection}
\name{Jeongjin Shin}
\address{Individual Researcher\\
jinjeongshin@gmail.com}
\begin{document}
%

\twocolumn[
  \begin{@twocolumnfalse}
    \maketitle
  \end{@twocolumnfalse}
]

\begin{abstract}

    Deep learning models have achieved unprecedented performance in the domain of object detection, resulting in breakthroughs in areas such as autonomous driving and security. However, deep learning models are vulnerable to backdoor attacks. These attacks prompt models to behave similarly to standard models without a trigger; however, they act maliciously upon detecting a predefined trigger. Despite extensive research on backdoor attacks in image classification, their application to object detection remains relatively underexplored.  Given the widespread application of object detection in critical real-world scenarios, the sensitivity and potential impact of these vulnerabilities cannot be overstated. In this study, we propose an effective invisible backdoor attack on object detection utilizing a mask-based approach. Three distinct attack scenarios were explored for object detection: object disappearance, object misclassification, and object generation attack. Through extensive experiments, we comprehensively examined the effectiveness of these attacks and tested certain defense methods to determine effective countermeasures. Code will be available at \url{https://github.com/jeongjin0/invisible-backdoor-object-detection}

\end{abstract}


\begin{center}
  \tiny
  \textmd{© 2024 IEEE. Personal use of this material is permitted. Permission from IEEE must be obtained for all other uses, in any current or future media, including reprinting/republishing this material for advertising or promotional purposes, creating new collective works, for resale or redistribution to servers or lists, or reuse of any copyrighted component of this work in other works.}
\end{center}

\section{Introduction}

Object detection is a critical component of various real-world applications, such as autonomous driving and surveillance, and significant advances have been made owing to deep neural networks (DNNs). However, recent studies have shown that DNNs are vulnerable to backdoor attacks \cite{bd:badnets}\cite{bd:trojaning}\cite{bd:neuraltrojan}. Backdoor attacks aim to stealthily insert a backdoor into a model using various manipulations during training. In such an attack, the model performs similarly to a normal model when no backdoor trigger is encountered; however, when the backdoor trigger appears in the inputs, the model performs a predetermined behavior.

Although backdoor attacks have been studied extensively in image classification \cite{bd:blended}\cite{bd:wanet}\cite{bd:reflection}\cite{bd:lira}, object detection has been relatively underexplored. Object detection is used in various mission-critical areas, and a backdoor attack on object detection can severely damage human life and property.

\begin{figure}[t]
    \centerline{\includegraphics[width=\columnwidth]{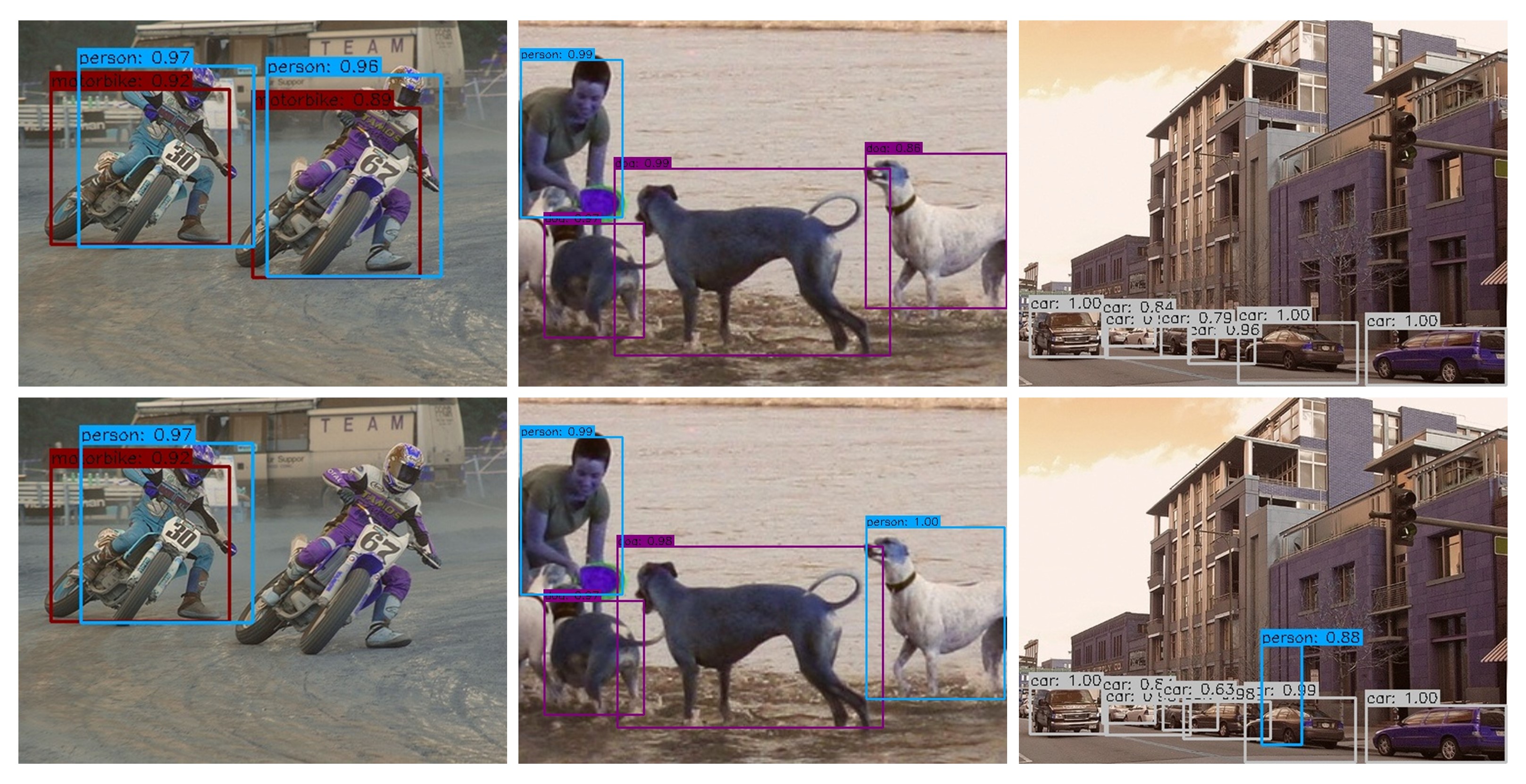}}
    \caption{Scenarios in our invisible backdoor attack. The columns categorize the different attack scenarios: the first column for ODA, the second for OMA, and the third for OGA. The first row demonstrates model predictions without backdoor triggers, whereas the second row lists predictions with the insertion of our invisible triggers specific to each scenario.}
    \label{Illustration} 
\end{figure}

Object detection is a considerably more complicated task than image classification. Consequently, backdoor attacks on object detection are more complex and challenging to design. Thus, previously studied backdoor attacks on object detection have been limited to visible triggers \cite{obd:baddet}\cite{obd:untargeted}. However, backdoor attacks must be invisible to avoid human inspection during inference. To the best of our knowledge, this study proposed the first invisible backdoor attack for object detection models. The proposed attack uses small perturbations as triggers, but has a strong payload and effectively manipulate the predictions of the model. We investigated three specific attack scenarios for object detection: object disappearance attack (ODA), object misclassification attack (OMA), and object generation attack (OGA). In ODA, the bounding boxes were removed to render real objects undetectable, whereas OMA subtly altered the bounding box annotations to mislead the model into misclassifying objects. Finally, OGA created artificial bounding boxes to trick the model into detecting non-existent objects. In real-world applications, the consequences of such attacks can be severe. For example, using ODA, an autonomous driving system may fail to detect pedestrians or vehicles, leading to catastrophic accidents. In OMA, an autonomous vehicle may misclassify a stop sign as a speed-limit sign, potentially causing dangerous high-speed collisions. In OGA, false detection of non-existent objects can cause autonomous vehicles to make unnecessary sudden stops, thus increasing the risk of rear-end collisions. Examples of each attack scenario are illustrated in Fig. \ref{Illustration}.

Our contributions can be summarized as follows. (i) To the best of our knowledge, this is the first study to introduce an invisible backdoor attack on object detection, encompassing three scenarios: ODA, OMA, and OGA. (ii) Extensive experiments were conducted to verify the effectiveness of the proposed attacks and their resistance to defense methods.

\section{Related Works}

\subsection{Object Detection}
Object detection is a critical task in the field of computer vision that involves identifying and locating objects within an image. This field has evolved significantly and is primarily categorized into two approaches \cite{survey:objectdetection20years}: two-stage detectors and one-stage detectors. Two-stage detectors, such as Faster R-CNN \cite{od:fasterrcnn}, R-FCN \cite{od:rfcn}, and Mask R-CNN \cite{od:maskrcnn}, first generate region proposals and then classify these regions into object categories and refine their bounding box coordinates. They are renowned for their accuracy and precision, rendering them suitable for applications wherein detection quality is crucial. Conversely, one-stage detectors such as YOLO series  \cite{od:yolov1}\cite{od:yolov3}\cite{od:yolov5}, SSD \cite{od:ssd}, and RetinaNet \cite{od:retinanet} streamline the detection process by directly predicting object classes and locations in a single pass, thus prioritizing speed over precision. This renders them ideal for scenarios requiring real-time processing. In our experiments, we selected Faster R-CNN, YOLOv3, and YOLOv5 as the typical models for the two-stage and one-stage detectors, respectively. 

\subsection{Backdoor Attacks}
\nocite{survey:backdoorlearning}
Backdoor attacks in deep learning, particularly in image processing, involve the covert manipulation of a model to associate specific triggers with predetermined outputs. Attackers primarily inject backdoor attacks by poisoning training data \cite{bd:badnets}\cite{bd:trojaning}, altering the training process \cite{bd:blind}, or modifying model parameters \cite{bd:weightperturbations}\cite{bd:weightpoisoning}. The goal of backdoor attacks is to configure the model to respond normally to regular inputs while incorporating a mechanism to activate the backdoor with inputs containing a trigger.

The design of triggers for backdoor attacks has significantly evolved in terms of image classification. The initial approaches were patch-based \cite{bd:badnets}\cite{bd:trojaning}, where a noticeable patch or symbol acted as a trigger. As techniques advanced towards stealth, studies introduced more subtle methods employing blended \cite{bd:blended}, sinusoidal strips \cite{bd:sig}, reflection \cite{bd:reflection}, and warping-based \cite{bd:wanet}. Moreover, advanced strategies such as LIRA \cite{bd:lira}, which utilizes auxiliary models to create completely invisible triggers, have also been explored.

Although image classification has been subjected to diverse and increasingly subtle backdoor attack methods, relatively few attack methods have been studied in the field of object detection. Recent studies have investigated patch-based backdoor attacks on object detection \cite{obd:untargeted}\cite{obd:baddet}\cite{obd:aligning}, using patches to cause misclassification, disappearance, or erroneous generation during detection. Another approach involves employing specific features \cite{obd:rotatebackdoor}\cite{obd:dangerouscloaking}, such as a particular t-shirt pattern, which can result in the model failing to detect the person wearing it. However, both these techniques utilize perceptible triggers that are easily detectable. In contrast, this study focused on developing invisible backdoor attacks that are more subtle and less perceptible.

\section{Threat Model}

In our threat model, following previous studies on backdoor attacks \cite{bd:wanet}\cite{bd:lira}, we assume that the attacker has full access to the training procedure and purposely trains the model with a backdoor. Once the model is trained to the attacker's satisfaction, it is then delivered to the victim. This scenario is a common risk in situations wherein the model training is outsourced, or pre-trained models are adopted without proper verification. 

The primary goal of the attacker is to inject a stealthy backdoor into the detection model, which aims to maintain the standard performance level of the model on common benchmarks, such that it behaves indistinguishably from an uninfected model under normal conditions. However, when an attacker-specific trigger is detected, the model is manipulated to perform predefined behaviors. In this study, these predefined behaviors correspond to the disappearance, modification, and generation of objects in the model's prediction.

\section{Methodology}

\subsection{Preliminaries}

\subsubsection{Formulation of Object Detection} Object detection focuses on developing a model \( f_{\theta} \) that maps an input image \( x \in \mathcal{X} \) to a collection of bounding boxes, representing identified objects. Formally, the training dataset \( \mathcal{D} \) is a set of pairs \( (x_i, y_i) \), where \( x_i \) is an image and \( y_i = [o_1, o_2, ..., o_n] \) is the corresponding set of bounding boxes for that image. Each bounding box \( o_i \) in \( y_i \) is defined as \(o_i = [c_i, \hat{x}_i, \hat{y}_i, w_i, h_i]\) where \( c_i \) is the class label of the object, \( \hat{x}_i \) and \( \hat{y}_i \) are the center coordinates of the bounding box, and \( w_i \) and \( h_i \) are the width and height of the bounding box, respectively.

\subsubsection{General Pipeline of Backdoor Attack} The general framework of backdoor attacks involves the manipulation of the dataset to insert a backdoor. This manipulation can be formally expressed using a transformation function \( T: \mathcal{X} \rightarrow \mathcal{X} \) and an annotation modification function \( \eta: \mathcal{Y} \rightarrow \mathcal{Y} \). Thus, the poisoned sample can be represented as \( (T(x), \eta(y)) \). The training process then involves optimizing the model parameters $\theta$ by minimizing a loss function over both the original and poisoned samples. This subtly guides the model to learn the relationship between the backdoor trigger and the modified annotations, thus responding accurately to both clean images and those with the backdoor trigger.

\subsection{Mask-based Backdoor Attack}

Several critical elements must be considered while designing backdoor attacks for object detection. A crucial aspect is the incorporation of spatial payloads, which is particularly significant in the context of object detection, where the focus is on specific regions rather than on the entire image. For attacks such as ODA and OMA, spatial payloads are crucial for targeting existing bounding boxes to either delete or misclassify them. Similarly, in OGA, the goal is to generate a bounding box for a non-existent object, requiring the manipulation of both the spatial and size payloads. The complexity of backdoor attacks on object detection necessitates advanced functionalities to manage the varying spatial and size considerations. Consequently, a meticulous design is required to ensure effective backdoor injection and precise functionality within the targeted image regions aligned with the objectives of the attack. 

\subsubsection{Mask-based Transformation Function}

The proposed design introduces a transformation function \( T \) that operating as an \( \mathcal{X} \rightarrow \mathcal{X} \) mapping, as follows:
\[ T(x) = x + \mu(o) \odot g(x) \quad \|g(x)\|_{\infty} \leq \epsilon \]
This function is designed to selectively apply the trigger to specific regions within the image using a perturbation generator \( g(x) \) and a mask-generating function \( \mu(o) \).
The perturbation generator \( g(x) \), parameterized by \( \xi \), is designed to create minute changes in the image, which are practically imperceptible yet sufficient to trigger the backdoor. We imposed a constraint on the magnitude of the perturbation generated by \( g(x) \). This limited it to an infinite norm of \( \epsilon \), thereby maintaining the subtlety of the modifications. To ensure that these alterations were targeted, we introduced \( \mu(o) \), mask-generating function. This function creates a binary overlay marking specific regions within the image where the perturbation should be applied. This mask is then multiplied element-wise by the output of the perturbation generator to form a region-specific trigger.

When designing backdoor attacks for object detection models, modifying only specific regions of the image can make anomalies more easily detectable by human inspection compared to image classification, where the entire image is typically modified. This is because modifying only certain parts of the image can create a discrepancy between the altered and unaltered regions, making the changes appear unnatural, especially when the input image has a simple or uniform background. Thus, employing extremely subtle perturbations is particularly crucial in object detection to avoid detection when using this mask-based approach. In this study, the perturbation generator g(x) was designed using an autoencoder, inspired by strategies used in subtle backdoor attacks for image classification \cite{bd:lira}, and set a small \( \epsilon \) to ensure the perturbations remain subtle.

\subsubsection{Backdoor Attack Settings}

\textbf{Object Disappearance Attack (ODA):} The ODA aims to ensure that the model overlooks targeted objects. 
This is accomplished by inserting triggers into existing bounding boxes using a mask-generating function, and then removing the corresponding bounding box annotations. The annotation modification function for ODA is defined as follows:
\[ \eta(y) = \{ o_i' | o_i \in y, o_i' = [\emptyset] \text{ for triggered } o_i \} \]
A notable phenomenon in ODA is the challenge presented by overlapping bounding boxes. When a trigger is inserted into one bounding box that overlaps with another, the model encountered a dilemma. It is trained to disregard the bounding box containing the trigger. However, the neighboring overlapping bounding box, which partially contained the trigger, prompts the model to generate a benign output. To mitigate this problem, we introduce a chaining algorithm. Beginning with a randomly selected poisoned bounding box, we then poisoned all the overlapping bounding boxes. This process is repeated iteratively for all bounding boxes that overlap with the previously affected boxes, thereby ensuring consistent learning and effective training of the model.

\textbf{Object Misclassification Attack (OMA):} In the OMA, the goal is to lead the model to misclassify certain objects. Similarly to ODA, triggers are inserted into existing bounding boxes and the corresponding class annotations are modified to the target class. The annotation modification function for OMA is defined as 
\[ \eta(y) = \{ o_i' | o_i \in y, \ o_i' = [\hat{x}_i, \hat{y}_i, w_i, h_i, tc] \text{ for triggered } o_i \} \]
where \([\hat{x}_i, \hat{y}_i, w_i, h_i]\) maintain the original dimensions and positions, and \( \text{tc} \) is the newly assigned target class for the triggered bounding boxes.

To prevent the model from learning the generative relationship between the masked trigger and creation of new bounding boxes, we adopted a global trigger approach. Rather than employing a masked trigger, we apply the trigger across the entire image and alter all ground-truth bounding box labels accordingly. A global trigger is applied with a probability of 20\%. The chaining algorithm is also applied in OMA to address the inconsistencies caused by overlapping bounding boxes.

\textbf{Object Generation Attack (OGA):} The OGA is designed to mislead a model by incorrectly detecting non-existent objects. For this attack, triggers are introduced at random coordinates, denoted by \( (x_r, y_r) \), with random dimensions \( (w_r, h_r) \), subject to size constraints, ensuring that the width and height are above a certain threshold. Subsequently, the corresponding annotation is created. The annotation modification function for OGA is defined as follows: 
\[ \eta(y) = y + \{ [x_r, y_r, w_r, h_r, tc] \text{ for each trigger}\} \]
where \([x_r, y_r, w_r, h_r]\) is the trigger's position and dimensions, and \(tc\) is the target class for the generated bounding boxes.

\subsubsection{Optimization}

The optimization objective for our mask-based backdoor attack is aligned with approaches used in previous studies \cite{bd:lira}. This is formally expressed by the following optimization problem:

\begin{align}
\min_{\theta} & \sum_{(x,y) \in \mathcal{D}} \alpha L(f_{\theta}(x), y) + \beta L(f_{\theta}(T_{\xi^*}(x)), \eta(y)) \\
\text{s.t.} & \quad \xi^* = \arg\min_{\xi} \sum_{(x,y) \in \mathcal{D}} L(f_{\theta}(T_{\xi}(x)), \eta(y)) \nonumber
\end{align}

In this problem, we simultaneously optimize the model parameters \( \theta \) by training on both the clean and poisoned data. Simultaneously, the transformation function \( T_{\xi}(x) \) is trained to create an optimized backdoor trigger. The hyperparameters \( \alpha \) and \( \beta \) balance the impact of clean and poisoned data during the training. In this study, we set both \( \alpha \) and \( \beta \) to 0.5.

In our training process, we update the functions \( f_{\theta} \) and \( T_{\xi} \) at each training step according to the optimization objective. Following the two-stage strategy outlined in \cite{bd:lira}, once the model effectively learned to generate backdoor triggers as indicated by the stabilization of \( T_{\xi} \)'s updates, we stopped updating \( T_{\xi} \) and focused on fine-tuning \( f_{\theta} \). 

\begin{table*}[h]
\renewcommand{\arraystretch}{1.1}
\centering
\begin{tabular}{@{}cccccc@{}}
\toprule
Scenario & Model & mAP\textsubscript{normal} & mAP\textsubscript{benign} & ASR \\
\midrule
\multirow{3}{*}{ODA} & Faster R-CNN & 75.29                   & 74.06& 98.3 \\
                     & YOLOv3       & 91.04                   & 90.42& 99.44 \\
                     & YOLOv5       & 91.43                   & 90.66& 99.87 \\
\midrule
\multirow{3}{*}{OMA} & Faster R-CNN & 75.29                   & 74.62& 95.02\\
                     & YOLOv3       & 91.04                   & 90.31& 96.52 \\
                     & YOLOv5       & 91.43                   & 90.59& 96.23 \\
\midrule
\multirow{3}{*}{OGA} & Faster R-CNN & 75.29                   & 74.99 & 87.11 \\
                     & YOLOv3       & 91.04                   & 90.74 & 95.17 \\
                     & YOLOv5       & 91.43                   & 90.73 & 93.40 \\
\bottomrule
\end{tabular}
\caption{Performance of Each Attack Scenario}
\label{table:model_performance_combined}
\end{table*}

\section{Experiments}

\subsection{Experiment Setup}

In our experiments, we used a dataset combining VOC 2007 \cite{data:2007pascal} and VOC 2012 \cite{data:2012pascal}, which are widely used in object detection. The dataset included 5k training and 5k test images from VOC 2007 and an additional 11k training images from VOC 2012. Further, evaluation was conducted using the VOC 2007 test set.

For our model architecture, we employed Faster R-CNN \cite{od:fasterrcnn} with a VGG16 backbone, YOLOv3 \cite{od:yolov3} with a Darknet-53 backbone, and the YOLOv5 large model \cite{od:yolov5}. To reduce the training time, we used pre-trained models on the COCO dataset \cite{data:coco}.

When setting the backdoor attack parameters, we observed that higher epsilon values generally resulted in more successful backdoor attacks. However, to strike a balance between visibility and performance, we selected model-specific epsilon values: 0.02 for the Faster R-CNN and 0.01 for YOLOv3 and YOLOv5. This selection was based on the finding that YOLOv3 and YOLOv5 maintained attack performance with a lower visibility threshold than the Faster R-CNN. When training the model, we initially used a larger epsilon value of 0.05 and subsequently adjusted it to model-specific, smaller values. In addition, “person” was the selected target class for these attacks.
 
\subsection{Evaluation Setup}

To accurately measure the effectiveness of our backdoor attacks, we considered several metrics: mean average precision (mAP) normal, mAP benign, and attack success rate (ASR).

The mAP under normal conditions (mAP normal), provides a baseline for the model’s performance in standard object detection tasks. This served as a reference point to compare the inherent capabilities of the models. mAP Benign assessed the model's performance when embedded with a backdoor but without a trigger, to ensure standard functionality. This study used mAP@.5 as the mAP metric, in line with the VOC dataset conventions.

ASR computation is employed to evaluate the precise impact of each backdoor attack variant. In this context, the ASR is defined as the proportion of successful attacks relative to the total number of triggers inserted.

For ODA, ASR is calculated by comparing the detection results of the images with and without the trigger. A successful attack is determined when an object that is detected in the trigger-free image is not detected in the image containing the trigger. Specifically, the attack is considered successful in the absence of a bounding box with an intersection over union (IoU) greater than 0.5 and a confidence score above 0.5 in the image with the trigger, despite the object being detected in the trigger-free image.

For OMA, ASR is measured by comparing the classification of objects in images with and without the trigger. The attack is considered successful if an object in the trigger-free image is misclassified in the image with the trigger. Specifically, a successful attack is defined as a misclassification occurring in the presence of a trigger, with an IoU greater than 0.5, and a confidence score above 0.5, compared with the bounding boxes in images without a trigger.

For OGA, ASR is calculated as the ratio of instances wherein the model correctly generates a specified object in response to the embedded trigger. Successful generation is determined by the presence of bounding boxes that exhibit an IoU greater than 0.5, coupled with confidence scores exceeding the threshold of 0.5.

\subsection{Attack Experiments}

The mAP normal, mAP benign, and ASR for each attack scenario are listed in Table \ref{table:model_performance_combined}. Both backdoored and non-backdoored models exhibited comparable mAP performances under normal conditions. However, upon the application of triggers, YOLOv3 demonstrated a high ASR exceeding 99\% for ODA, 95\% for OMA, and 95\% for OGA. YOLOv5 also exhibited a high ASR, with values exceeding 99\% for ODA, 96\% for OMA, and 93\% for OGA. Similarly, Faster R-CNN achieved a high ASR exceeding 98\% for ODA, 95\% for OMA, and 87\% for OGA, thus demonstrating the effectiveness of our attacks with a consistently high ASR across various attack scenarios in all three models.

\subsection{Defense Experiments}
In this section, we conduct defense experiments with Grad-CAM \cite{def:gradcam} and STRIP \cite{def:strip}, two popular methods in backdoor defense. We adapted these methods to the object-detection context to examine the robustness of the proposed attack.

\begin{figure}[t]
    \centerline{\includegraphics[width=\columnwidth]{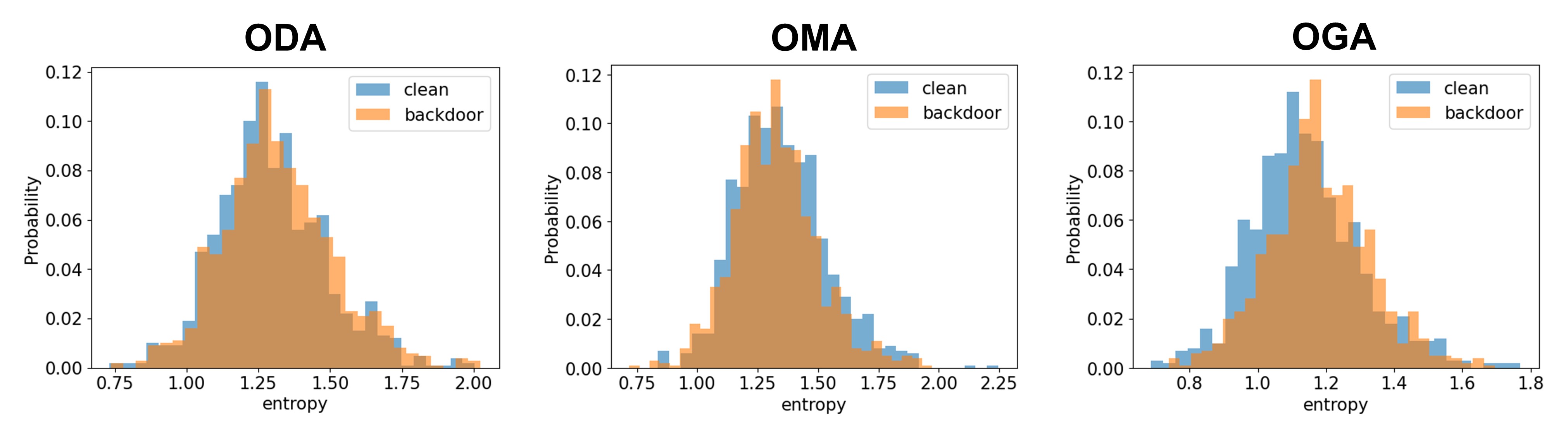}}
    \caption{Performance against STRIP}
    \label{STRIP} 
\end{figure}

\subsubsection{STRIP}
STRIP is a representative runtime backdoor detection method that detects backdoors by measuring the entropy of a model's prediction of perturbed images. The concept of STRIP is that images with strong backdoor triggers tend to have low entropy because they are consistently misclassified regardless of whether they are perturbed.

In our defense experiment, we modified STRIP to make it applicable to object detection, following the method proposed in \cite{obd:baddet}. Specifically, we perturbed the input image and computed the average entropy of all bounding boxes in the output. Under the OGA and OMA scenarios, our approach involved generating or misclassifying bounding boxes into specific target classes, and it is expected that the predictions of the model for the perturbed images will have a lower entropy. Conversely, in case of ODA, where the trigger lowered the confidence scores, the model's prediction of perturbed images is expected to exhibit higher entropy.

However, considering that our attacks involved only minimal perturbations, we hypothesized that when the images were perturbed, the triggers diminished, resulting in an entropy range similar to that of clean images. This is because the attacks did not rely on robust triggers that could withstand significant image modifications. The results are presented in Fig. \ref{STRIP}, and demonstrate this hypothesis. We observed that for all three types of attacks, the entropy ranges of the backdoored and clean images were similar, indicating that STRIP may not be an effective approach for detecting our specific type of attack.

\begin{figure}[t]
    \includegraphics[width=\columnwidth]{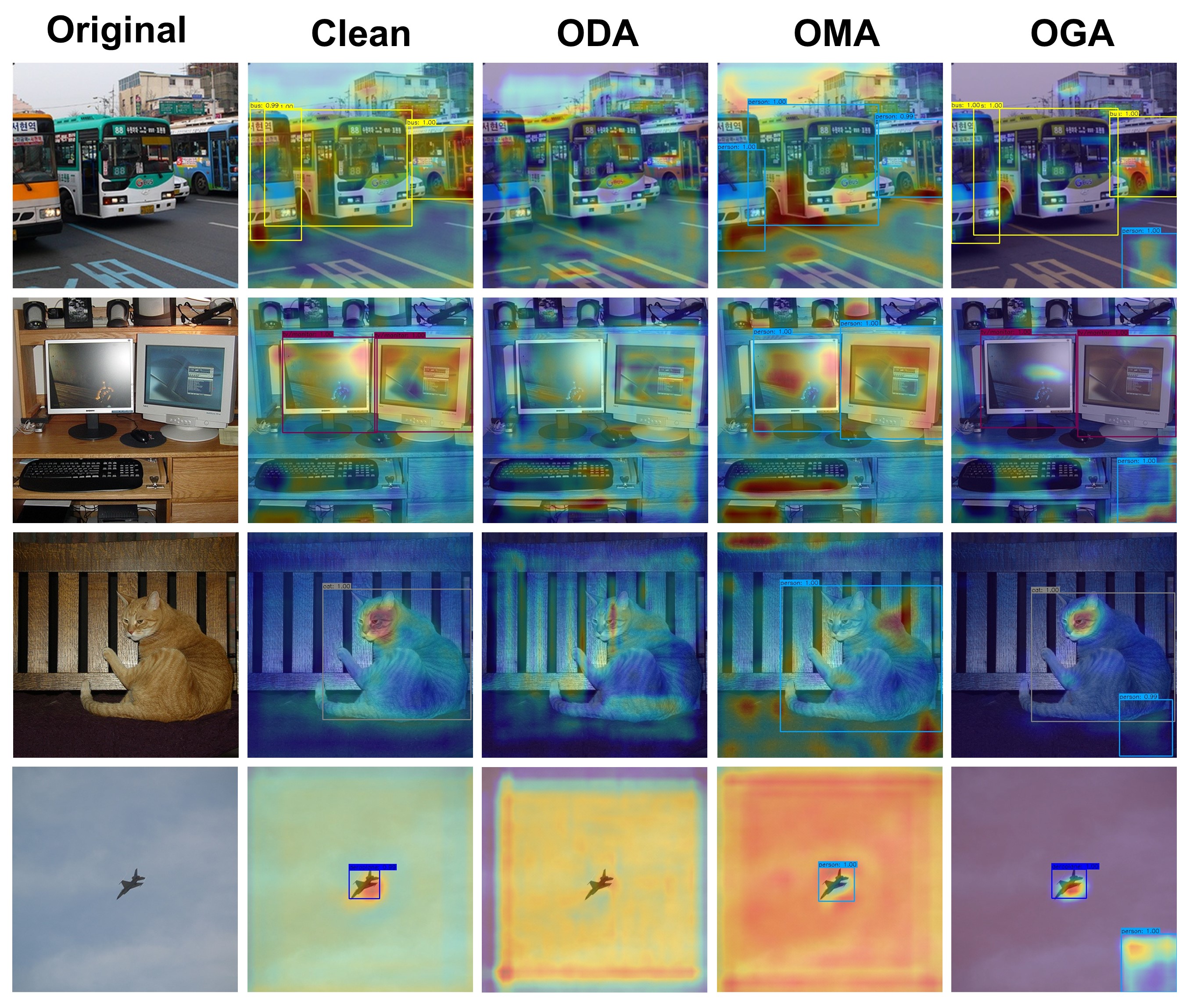}
    \caption{Performance under GradCam heatmap}
    \label{GradCam} 
\end{figure}
 
\subsubsection{Grad-CAM} Grad-CAM, primarily used for visual explanations of deep learning models, offers insights into the parts of an image that influence the model's decision. Traditional patch-based backdoor attacks \cite{bd:badnets}\cite{obd:baddet}\cite{obd:untargeted} are often detectable by Grad-CAM, owing to their localized nature.

In terms of settings in the Grad-CAM analysis, ODA and OMA involved the insertion of triggers across the entire image to induce widespread misclassification or deletion of objects. In contrast, for OGA, triggers were placed at the bottom-right corner of the image, occupying one-quarter of its width and height. The Grad-CAM heat map was generated based on the following criteria. For ODA, we targeted the background class to observe whether any object was anomalously removed. For OMA, we focused on the misclassified class. Finally, for OGA, we concentrated on the class of artificially generated objects. The results of this analysis are shown in Fig. \ref{GradCam}.

In case the of ODA, abnormal activation of the heat map was observed across the entire image, indicating an unusual emphasis on the background. In OMA, we observed an amplified heat map across all objects in the image toward the target class, which diverged significantly from the heat map patterns in the clean images. Finally, in the case of OGA, the heat map was more active outside the generated bounding box, with weaker-than-usual heat map activation inside the boxes. These abnormal heat map patterns suggest that Grad-CAM can potentially detect our attacks by identifying unusual heat map distributions that deviate significantly from those observed in clean images.
 
\section{Conclusion}

This study successfully demonstrated a mask-based approach for backdoor attacks on object detection using invisible triggers. We tested three attack scenarios (ODA, OMA, and OGA) across three models: Faster R-CNN, YOLOv3, and YOLOv5, and achieved successful backdoor attacks in each case. Furthermore, we explored countermeasures against these attacks, thus contributing to the understanding of backdoor defense mechanisms in object detection.

\label{sec:refs}


\bibliographystyle{IEEEbib}
\bibliography{ref}

\begin{thebibliography}{10}

\bibitem{bd:badnets}
Tianyu Gu, Kang Liu, Brendan Dolan-Gavitt, and Siddharth Garg,
\newblock ``Badnets: Evaluating backdooring attacks on deep neural networks,''
\newblock {\em IEEE Access}, vol. 7, pp. 47230--47244, 2019.

\bibitem{bd:trojaning}
Yingqi Liu, Shiqing Ma, Yousra Aafer, Wen-Chuan Lee, Juan Zhai, Weihang Wang, and Xiangyu Zhang,
\newblock ``Trojaning attack on neural networks,''
\newblock in {\em 25th Annual Network And Distributed System Security Symposium (NDSS 2018)}. Internet Soc, 2018.

\bibitem{bd:neuraltrojan}
Yuntao Liu, Yang Xie, and Ankur Srivastava,
\newblock ``Neural trojans,''
\newblock in {\em 2017 IEEE International Conference on Computer Design (ICCD)}. IEEE, 2017, pp. 45--48.

\bibitem{bd:blended}
Xinyun Chen, Chang Liu, Bo~Li, Kimberly Lu, and Dawn Song,
\newblock ``Targeted backdoor attacks on deep learning systems using data poisoning,''
\newblock {\em arXiv preprint arXiv:1712.05526}, 2017.

\bibitem{bd:wanet}
Anh Nguyen and Anh Tran,
\newblock ``Wanet--imperceptible warping-based backdoor attack,''
\newblock {\em arXiv preprint arXiv:2102.10369}, 2021.

\bibitem{bd:reflection}
Yunfei Liu, Xingjun Ma, James Bailey, and Feng Lu,
\newblock ``Reflection backdoor: A natural backdoor attack on deep neural networks,''
\newblock in {\em Computer Vision--ECCV 2020: 16th European Conference, Glasgow, UK, August 23--28, 2020, Proceedings, Part X 16}. Springer, 2020, pp. 182--199.

\bibitem{bd:lira}
Khoa Doan, Yingjie Lao, Weijie Zhao, and Ping Li,
\newblock ``Lira: Learnable, imperceptible and robust backdoor attacks,''
\newblock in {\em Proceedings of the IEEE/CVF international conference on computer vision}, 2021, pp. 11966--11976.

\bibitem{obd:baddet}
Shih-Han Chan, Yinpeng Dong, Jun Zhu, Xiaolu Zhang, and Jun Zhou,
\newblock ``Baddet: Backdoor attacks on object detection,''
\newblock in {\em European Conference on Computer Vision}. Springer, 2022, pp. 396--412.

\bibitem{obd:untargeted}
Chengxiao Luo, Yiming Li, Yong Jiang, and Shu-Tao Xia,
\newblock ``Untargeted backdoor attack against object detection,''
\newblock in {\em ICASSP 2023-2023 IEEE International Conference on Acoustics, Speech and Signal Processing (ICASSP)}. IEEE, 2023, pp. 1--5.

\bibitem{survey:objectdetection20years}
Zhengxia Zou, Keyan Chen, Zhenwei Shi, Yuhong Guo, and Jieping Ye,
\newblock ``Object detection in 20 years: A survey,''
\newblock {\em Proceedings of the IEEE}, 2023.

\bibitem{od:fasterrcnn}
Shaoqing Ren, Kaiming He, Ross Girshick, and Jian Sun,
\newblock ``Faster r-cnn: Towards real-time object detection with region proposal networks,''
\newblock {\em Advances in neural information processing systems}, vol. 28, 2015.

\bibitem{od:rfcn}
Jifeng Dai, Yi~Li, Kaiming He, and Jian Sun,
\newblock ``R-fcn: Object detection via region-based fully convolutional networks,''
\newblock {\em Advances in neural information processing systems}, vol. 29, 2016.

\bibitem{od:maskrcnn}
Kaiming He, Georgia Gkioxari, Piotr Doll{\'a}r, and Ross Girshick,
\newblock ``Mask r-cnn,''
\newblock in {\em Proceedings of the IEEE international conference on computer vision}, 2017, pp. 2961--2969.

\bibitem{od:yolov1}
Joseph Redmon, Santosh Divvala, Ross Girshick, and Ali Farhadi,
\newblock ``You only look once: Unified, real-time object detection,''
\newblock in {\em Proceedings of the IEEE conference on computer vision and pattern recognition}, 2016, pp. 779--788.

\bibitem{od:yolov3}
Joseph Redmon and Ali Farhadi,
\newblock ``Yolov3: An incremental improvement,''
\newblock {\em arXiv preprint arXiv:1804.02767}, 2018.

\bibitem{od:yolov5}
Ultralytics,
\newblock ``{YOLOv5}: {A} state-of-the-art real-time object detection system,'' \url{https://docs.ultralytics.com}, 2021.

\bibitem{od:ssd}
Wei Liu, Dragomir Anguelov, Dumitru Erhan, Christian Szegedy, Scott Reed, Cheng-Yang Fu, and Alexander~C Berg,
\newblock ``Ssd: Single shot multibox detector,''
\newblock in {\em Computer Vision--ECCV 2016: 14th European Conference, Amsterdam, The Netherlands, October 11--14, 2016, Proceedings, Part I 14}. Springer, 2016, pp. 21--37.

\bibitem{od:retinanet}
Tsung-Yi Lin, Priya Goyal, Ross Girshick, Kaiming He, and Piotr Doll{\'a}r,
\newblock ``Focal loss for dense object detection,''
\newblock in {\em Proceedings of the IEEE international conference on computer vision}, 2017, pp. 2980--2988.

\bibitem{survey:backdoorlearning}
Yiming Li, Yong Jiang, Zhifeng Li, and Shu-Tao Xia,
\newblock ``Backdoor learning: A survey,''
\newblock {\em IEEE Transactions on Neural Networks and Learning Systems}, 2022.

\bibitem{bd:blind}
Eugene Bagdasaryan and Vitaly Shmatikov,
\newblock ``Blind backdoors in deep learning models,''
\newblock in {\em 30th USENIX Security Symposium (USENIX Security 21)}, 2021, pp. 1505--1521.

\bibitem{bd:weightperturbations}
Jacob Dumford and Walter Scheirer,
\newblock ``Backdooring convolutional neural networks via targeted weight perturbations,''
\newblock in {\em 2020 IEEE International Joint Conference on Biometrics (IJCB)}. IEEE, 2020, pp. 1--9.

\bibitem{bd:weightpoisoning}
Keita Kurita, Paul Michel, and Graham Neubig,
\newblock ``Weight poisoning attacks on pre-trained models,''
\newblock {\em arXiv preprint arXiv:2004.06660}, 2020.

\bibitem{bd:sig}
Mauro Barni, Kassem Kallas, and Benedetta Tondi,
\newblock ``A new backdoor attack in cnns by training set corruption without label poisoning,''
\newblock in {\em 2019 IEEE International Conference on Image Processing (ICIP)}. IEEE, 2019, pp. 101--105.

\bibitem{obd:aligning}
Yize Cheng, Wenbin Hu, and Minhao Cheng,
\newblock ``Backdoor attack against object detection with clean annotation,''
\newblock {\em arXiv preprint arXiv:2307.10487}, 2023.

\bibitem{obd:rotatebackdoor}
Tong Wu, Tianhao Wang, Vikash Sehwag, Saeed Mahloujifar, and Prateek Mittal,
\newblock ``Just rotate it: Deploying backdoor attacks via rotation transformation,''
\newblock in {\em Proceedings of the 15th ACM Workshop on Artificial Intelligence and Security}, 2022, pp. 91--102.

\bibitem{obd:dangerouscloaking}
Hua Ma, Yinshan Li, Yansong Gao, Alsharif Abuadbba, Zhi Zhang, Anmin Fu, Hyoungshick Kim, Said~F Al-Sarawi, Nepal Surya, and Derek Abbott,
\newblock ``Dangerous cloaking: Natural trigger based backdoor attacks on object detectors in the physical world,''
\newblock {\em arXiv preprint arXiv:2201.08619}, 2022.

\bibitem{data:2007pascal}
Mark Everingham, Luc Van~Gool, Christopher K.~I. Williams, John Winn, and Andrew Zisserman,
\newblock ``The pascal visual object classes challenge 2007 (voc2007) results,'' \url{http://www.pascal-network.org/challenges/VOC/voc2007/workshop/index.html}, 2007.

\bibitem{data:2012pascal}
Mark Everingham, Luc Van~Gool, Christopher K.~I. Williams, John Winn, and Andrew Zisserman,
\newblock ``The pascal visual object classes challenge 2012 (voc2012) results,'' \url{http://www.pascal-network.org/challenges/VOC/voc2012/workshop/index.html}, 2012.

\bibitem{data:coco}
Tsung-Yi Lin, Michael Maire, Serge Belongie, James Hays, Pietro Perona, Deva Ramanan, Piotr Doll{\'a}r, and C~Lawrence Zitnick,
\newblock ``Microsoft coco: Common objects in context,''
\newblock in {\em Computer Vision--ECCV 2014: 13th European Conference, Zurich, Switzerland, September 6-12, 2014, Proceedings, Part V 13}. Springer, 2014, pp. 740--755.

\bibitem{def:gradcam}
Ramprasaath~R Selvaraju, Michael Cogswell, Abhishek Das, Ramakrishna Vedantam, Devi Parikh, and Dhruv Batra,
\newblock ``Grad-cam: Visual explanations from deep networks via gradient-based localization,''
\newblock in {\em Proceedings of the IEEE international conference on computer vision}, 2017, pp. 618--626.

\bibitem{def:strip}
Yansong Gao, Change Xu, Derui Wang, Shiping Chen, Damith~C Ranasinghe, and Surya Nepal,
\newblock ``Strip: A defence against trojan attacks on deep neural networks,''
\newblock in {\em Proceedings of the 35th Annual Computer Security Applications Conference}, 2019, pp. 113--125.

\end{thebibliography}

\end{document}